\documentclass{article}
\usepackage{spconf}
\usepackage{amsmath,graphicx}
\usepackage{color}
\usepackage{subcaption}
\usepackage{multirow}
\usepackage{booktabs}
\usepackage{lipsum}
\usepackage{xcolor}
\usepackage{csvsimple}
\usepackage{url}

\newcommand{\Tab}[1]{Tab.~\ref{#1}}
\newcommand{\Fig}[1]{Fig.~\ref{#1}}

\newcommand{\Sec}[1]{Section~\ref{#1}}

\title{Tabular Transformers for Modeling Multivariate Time Series}

\name{\parbox{\textwidth}{\centering
    Inkit Padhi, Yair Schiff, Igor Melnyk, Mattia Rigotti, Youssef Mroueh, Pierre Dognin\\ Jerret Ross, Ravi Nair, Erik Altman
  }
}
\address{IBM Research, T.J Watson Research Center  \&  MIT-IBM Watson AI Lab }
 
\begin{document}
%
\maketitle
\begin{abstract}
Tabular datasets are ubiquitous in data science applications. Given their importance, it seems natural to apply state-of-the-art deep learning algorithms in order to fully unlock their potential.
Here we propose neural network models that represent \emph{tabular time series} that can optionally leverage their hierarchical structure.
This results in two architectures for tabular time series: one for learning representations that is analogous to BERT and can be pre-trained end-to-end and used in downstream tasks, and one that is akin to GPT and can be used for generation of realistic synthetic tabular sequences.
We demonstrate our models on two datasets: a synthetic credit card transaction dataset, where the learned representations are used for fraud detection and synthetic data generation, and on a real pollution dataset, where the learned encodings are used to predict atmospheric pollutant concentrations. Code and data are available at \url{https://github.com/IBM/TabFormer}.
\end{abstract}

%
\section{Introduction}
\label{sec:intro}

Tabular datasets are ubiquitous across many industries, especially in vital sectors such as healthcare and finance.
Such industrial datasets often contain sensitive information, raising privacy and confidentiality issues that preclude their public release and limit their analysis to methods that are compatible with an appropriate anonymization process.
We can distinguish between two types of tabular data: \emph{static} tabular data that corresponds to independent rows in a table, and \emph{dynamic} tabular data that corresponds to tabular time series, also referred to also as multivariate time series.
The machine learning and deep learning communities have devoted considerable effort to learning from static tabular data, as well as generating synthetic static tabular data that can be released as a privacy compliant surrogate of the original data.
On the other hand, less effort has been devoted to the more challenging dynamic case, where it is important to also account for the temporal component of the data.
The purpose of this paper is to remedy this gap by proposing deep learning techniques to: 1) learn useful representation of tabular time series that can be used in downstream tasks such as classification or regression and 2) generate realistic synthetic tabular time series.  

\emph{Tabular time series} represent a hierarchical structure that we leverage by endowing transformer-based language models with field-level transformers, which encode individual rows into embeddings that are in turn treated as embedded tokens that are passed to BERT \cite{devlin2018bert}. 
This results in an alternative architectures for  tabular time series encoding that can be pre-trained end-to-end for representation learning that we call Tabular BERT (TabBERT). Another important contribution is adapting state-of-the-art (SOTA) language generative models GPT \cite{GPT} to produce realistic synthetic tabular data that we call Tabular GPT (TabGPT).
A key ingredient of our \emph{language metaphor} in modeling tabular time series is the quantization of continuous fields, so that each field is defined on a finite vocabulary, as in language modeling.

As mentioned, static tabular data have been widely analyzed in the past, typically with feature engineering and classical learning schemes such as gradient boosting or random forests.
Recently, \cite{arik2019tabnet} introduced TabNet, which uses attention to perform feature selection across fields and shows the advantages of deep learning over classical approaches. 
A more recent line of work \cite{herzig-etal-2020-tapas,yin20acl} concurrent to ours, deals with the joint processing of static tabular and textual data using transformer architectures, such as BERT, with the goal of querying tables with natural language. 
 These works consider the static case, and to the best of our knowledge, our work is the first to address tabular time series using transformers.

On the synthetic generation side, a plethora of work \cite{xu2020synthesizing,xu2018synthesizing,xu2019modeling,karlsson2020synthesis,camino2020working} are dedicated to generating static tabular data using Generative Adversarial Networks (GANs), conditional GANs, and variational Auto-Encoders.
\cite{assefa2020generating,efimov2020using} argue for the importance of synthetic generation on financial tabular data in order to preserve user privacy and to allow for training on cloud-based solutions without compromising real users' sensitive information.
Nevertheless, their generation scheme falls short of modeling the temporal dependency in the data. 
Our work addresses this crucial aspect in particular.  
In summary, the main contributions of our paper are:

\noindent $\bullet$ We propose Hierarchical Tabular BERT to learn representations of tabular time series that can be used in downstream tasks such as classification or regression. 

\noindent $\bullet$ We propose TabGPT to synthesize realistic tabular time series data.
 
\noindent $\bullet$ We train our proposed models on a synthetic credit card transaction dataset, where the learned encodings are used for a downstream fraud detection task and for synthetic data generation. 
We also showcase our method on a public real-world pollution dataset, where the learned encodings are used to predict the concentration of pollutant.
  
 \noindent $\bullet$  We open-source our synthetic credit-card transactions dataset  to encourage future research on this type of data.  The code to reproduce all experiments in this paper is available at \url{https://github.com/IBM/TabFormer}. Our code is built within HuggingFace's framework \cite{Wolf2019HuggingFacesTS}.

\section{TabBERT: Unsupervised Learning of Multivariate time series Representation}

\subsection{From  Language Modeling to Tabular Time Series }
\label{sec:about_data}
\begin{figure}[ht]
    \centering
    \includegraphics[width=1\linewidth]{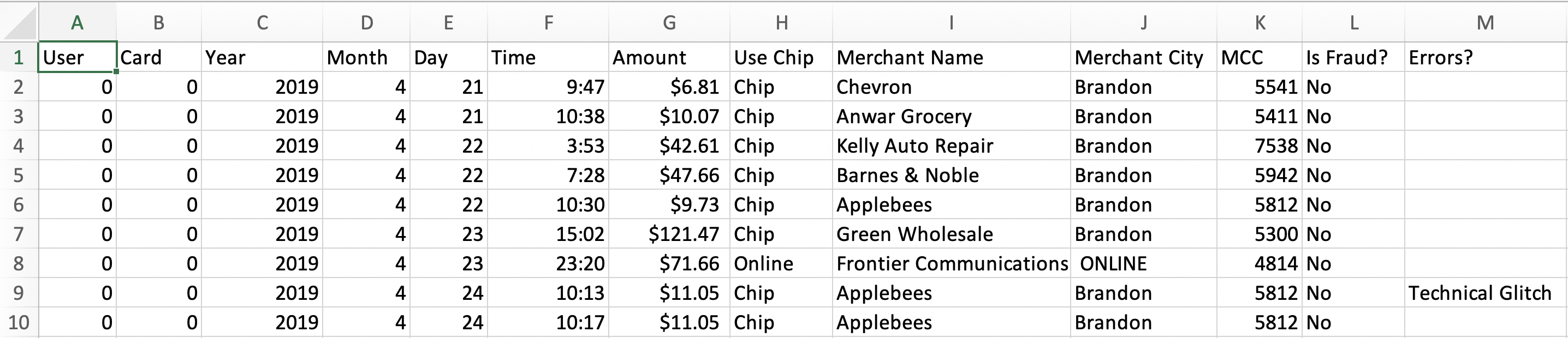}
    \caption{An example of sequential tabular data, where each row is a transaction. \textit{A} to \textit{M} are the fields of the transactions. Some of the fields are categorical, others are continuous, but through quantization we convert all fields into categorical. Each field is then processed to build its own local vocabulary. A single sample is defined as some number of contiguous transactions, for example rows 1 through 10, as shown in this figure.}
    \label{diag:tabdata}
    \vskip -0.1in
\end{figure}
In \Fig{diag:tabdata}, we give an example of tabular time series, that is a sequence of card transactions for a particular user. 
Each row consists of  fields that can be continuous or categorical. In order to unlock the potential of language modeling techniques for tabular data, we quantize continuous fields so that each field is defined on its own local finite vocabulary. We define a sample as a sequence of rows (transactions in this case). The main difference with NLP is that we have a sequence of structured rows consisting each of fields defined on local vocabulary.
As introduced in previous sections, unsupervised learning for multivariate time series representations require modeling both inter- and intra-transaction dependencies. In the next section, we show how to exploit this hierarchy in learning unsupervised representations for tabular time series.  

\subsection{Hierarchical Tabular BERT}

\begin{figure}[t!]
    \centering
        \includegraphics[width=\linewidth]{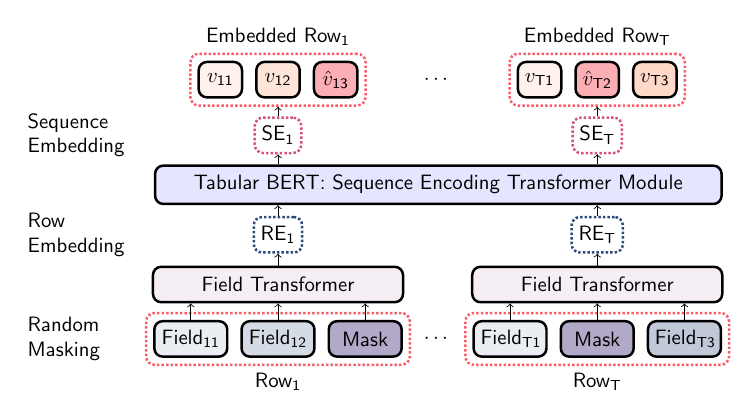}
    \caption{TabBERT: Field level masking and cross entropy.}
    \label{diag:tabbert}
    \vskip -0.1in
\end{figure}

In order to learn representations for multivariate tabular data, we use a recipe similar to the one employed for training language representation using BERT. The encoder is trained through masked language modeling (MLM), i.e by predicting masked tokens.
More formally, given a table with $M$ rows and $N$ columns (or fields), an
input, $t$, to TabBERT is represented as a windowed series of $T$ time-dependent rows,
\begin{align}
t = [t_{i+1}, t_{i+2}, ..., t_{i+T}],
\end{align}
where $T$ (\(\ll\)$M$) is the number of consecutive rows, and selected with a window offset (or stride). 
TabBERT is a variant of BERT, which accommodates the temporal nature of rows in the tabular input. 
As shown in \Fig{diag:tabbert}, TabBERT encodes the series of transactions in a hierarchical fashion. 
The field transformer processes rows individually, creating transaction/row embeddings. 
These transaction embeddings are then fed to a second-level transformer to create sequence embeddings. 
In other words, the field transformer tries to capture intra-transaction relationships (local), whereas the sequence transformer encodes inter-transaction relationships (global), capturing the temporal component of the data. Note that hierarchical transformer has already been proposed in NLP in order to hierarchically model documents \cite{pappagari2019hierarchical,zhang-etal-2019-hibert}.  

Many pre-trained transformer models on domain specific data have recently been successfully applied in various downstream tasks. More specifically, 
BioBERT \cite{lee2020biobert}, VideoBERT \cite{sun2019videobert}, ClinicalBERT \cite{huang2019clinicalbert} are pre-trained efficiently on various domains, such as biomedical, YouTube videos, and clinical electronic health record, respectively. Representations from these BERT variants  achieve SOTA results on  tasks ranging from video captioning to hospital readmission. 
In order to ascertain the richness of learned TabBERT representation, we study two downstream tasks: classification (\Sec{sec:classification}) and regression (\Sec{sec:regression}).
 
\subsection{TabBERT Features for Classification} 
\label{sec:classification}

\noindent \textbf{Transaction Dataset} One of the contributions of this work is in introducing a new synthetic corpus for credit card transactions. 
The transactions are created using a rule-based generator where values are produced by stochastic sampling techniques, similar to a method followed by \cite{altman2019synthesizing}. 
Due to privacy concerns, most of the existing public transaction datasets are either heavily anonymized or preserved through PCA-based transformations, and thus distort the real data distributions. 
The proposed dataset has 24 million transactions from 20,000 users. 
Each transaction (row) has 12 fields (columns) consisting of both continuous and discrete nominal attributes, such as merchant name, merchant address, transaction amount, etc. \\

\noindent \textbf{Training TabBERT}
For training TabBERT on our transaction dataset, we create samples as sliding windows of 10 transactions, with a stride of 5. 
The continuous and categorical values are quantized, and a vocabulary is created, as described in \Sec{sec:about_data}. 
Note that during training we exclude the label column \texttt{isFraud?}, to avoid biasing the learned representation for the downstream fraud detection task. 
Similar to strategies used by \cite{devlin2018bert}, we mask 15\% of a sample's fields, replacing them with the \textsc{[Mask]} token, and predict the original field token with cross entropy loss.\\

\noindent \textbf{Fraud Detection Task}
For fraud detection, we create samples by combining 10 contiguous rows (with a stride of 10) in a time-dependent manner for each user.
In total, there are 2.4M samples with 29,342 labeled as fraudulent.
Since in real world the fraudulent transactions are very rare events, a similar trend is observed in our synthetic data, resulting in an imbalanced, non-uniform distribution of fraudulent and non-fraudulent class labels. To account for this imbalance, during training, we upsample the fraudulent class to roughly equalize the frequencies of both classes. We evaluate performance of different methods using F1 binary score, on a test set consisting of 480K samples. 
As a baseline, we use a multi-layer perceptron (MLP) trained directly on the embeddings of the raw features. 
In order to model temporal dependencies, we also use an LSTM network baseline on the raw embedded features. 
In both  cases, we pool the encoder outputs at individual row level to create $E_i$ (see \Fig{diag:tabbert}) before doing classification. 
In \Tab{tab:result}, we compare the baselines and the methods based on TabBERT features while using the same architectures for the prediction head.
During MLP and LSTM networks training, with TabBERT as the feature extractor, we freeze the TabBERT network foregoing any update of its weights. 
As can be seen from the table, the inclusion of TabBERT features boosts the F1 for the fraud detection task.

\subsection{TabBERT Features for Regression Tasks}
\label{sec:regression}
 \noindent\textbf{Pollution Dataset} For the regression task, we use a public UCI dataset (Beijing PM2.5 Data) for predicting both PM2.5 and PM10 air concentration for 12 monitoring sites, each containing around 35k entries (rows). Every row has 11 fields with a mix of both continuous and discrete values. For a detailed description of the data, please refer to \cite{liang2015assessing}. 
 Similar to the pre-processing steps for our transaction dataset, we quantize the continuous features, remove the targets (\textit{PM2.5} and \textit{PM10}), and create samples by combining 10 time-dependent rows with a stride of 5. We use 45K samples for training and
  report a combined RMSE for both targets from the test set of 15K samples. 
 As reported in \Tab{tab:result}, using TabBERT features shows significant improvement in terms of RMSE over the case of using simple raw embedded features. This consistent performance gain when using TabBERT features for both classification and regression tasks underlines the richness of representations learned from TabBERT.
 

\begin{table}[t!]
\centering
\begin{tabular}{cccll}
\multicolumn{2}{l}{} & \multicolumn{2}{c}{\textbf{Fraud}} & \textbf{PRSA} \\
\multicolumn{1}{l}{\textbf{Features}} & \textbf{Prediction Head} & \multicolumn{2}{c}{F1} & \multicolumn{1}{c}{RMSE} \\ 
\hline
\multirow{2}{*}{Raw} & MLP & \multicolumn{2}{c}{0.74} & 38.5 \\
 & LSTM & \multicolumn{2}{c}{0.83} & 43.3 \\ \hline
\multirow{2}{*}{TabBERT} & MLP & \multicolumn{2}{c}{0.76} & 34.2 \\
 & LSTM & \multicolumn{2}{c}{\textbf{0.86}} & \textbf{32.8}
\end{tabular}%
\caption{Performance comparison on the classification task of fraud detection (Fraud), and the regression task of pollution prediction (PRSA) for two approaches: one based on TabBERT features and the baseline using raw data. We compare two architectures: MLP and LSTM for the downstream tasks.}
\label{tab:result}
\vskip -0.2in
\end{table}


\section{TabGPT: Generative modeling of Multivariate Time Series Tabular Data}\label{sec:tab_gpt}
Another useful application of language modeling in the context of tabular, time-series data is the preservation of data privacy.
GPT models trained on large corpora have demonstrated human-level capabilities in the domain of text generation.
In this work, we apply the generative capabilities of GPT as a proof of concept for creating synthetic tabular data that is close in distribution to the true data, with the advantage of not exposing any sensitive information.
Specifically, we train a GPT model (referred to throughout as TabGPT) on user-level data from the credit card dataset in order to generate synthetic transactions that mimic a user's purchasing behavior.
This synthetic data can subsequently be used in downstream tasks without the precautions that would typically be necessary when handling private information.

\begin{figure}
    \centering
        \includegraphics[width=\linewidth]{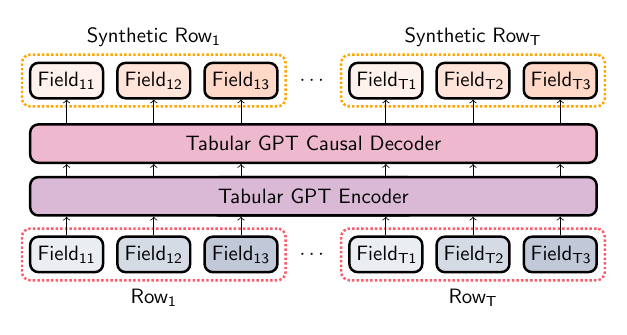}
    \caption{TabGPT: Synthetic Transaction GPT Generator.}
    \label{diag:gpt_diagram}
    \vskip -0.15in
\end{figure}

We begin, as with TabBERT, by quantizing the data to create a finite vocabulary for each field.
To train the TabGPT model, we select specific users from the dataset.
By ordering a user's transactions chronologically and segmenting them into sequences of ten transactions, the model learns to predict future behavior from past transactions, similar to how GPT language models are trained on text data to predict future tokens from past context.
We apply this approach to two of the users that have a relatively high volume of transactions, each with $\sim$60k transactions.
For each user, we train a separate TabGPT model, which is depicted in \Fig{diag:gpt_diagram}. 
Unlike with TabBERT, we do not employ the hierarchical structure of passing each field into a field-level transformer, but rather we pass sequences of transactions separated by a special \textsc{[Sep]} token directly to the GPT encoder network.

After training, synthetic data is generated by again segmenting a user's transaction data into sequences of ten transactions, passing the first transaction of each group of ten to the model, and predicting the remaining nine.
To evaluate a model's generative capabilities we examine how it captures both the aggregate and time-dependent features of the data.

The quantization of non-categorical data, which enables the use of a finite vocabulary for each field, renders field level evaluation of the fidelity of TabGPT to the real data more straightforward.
Namely, for each field, we compute and compare histograms for both ground truth and generated data on an aggregate level over all timestamps.
To measure proximity of the true and synthetic distributions we calculate the $\chi^2$ distance between histograms, defined as:
$\chi^2(\mathcal{X}, \mathcal{X'}) = \frac{1}{2}\sum_{i=1}^{n}\frac{(x_i - x'_i)^2}{(x_i + x'_i)}$
where $x_i, x'_i$ are columns from the corresponding transactions ($i=1..n$) from the true ($\mathcal{X}$) and generated ($\mathcal{X'}$) distributions, respectively.
\begin{figure}[ht]
    \centering
    \includegraphics[width=1\linewidth]{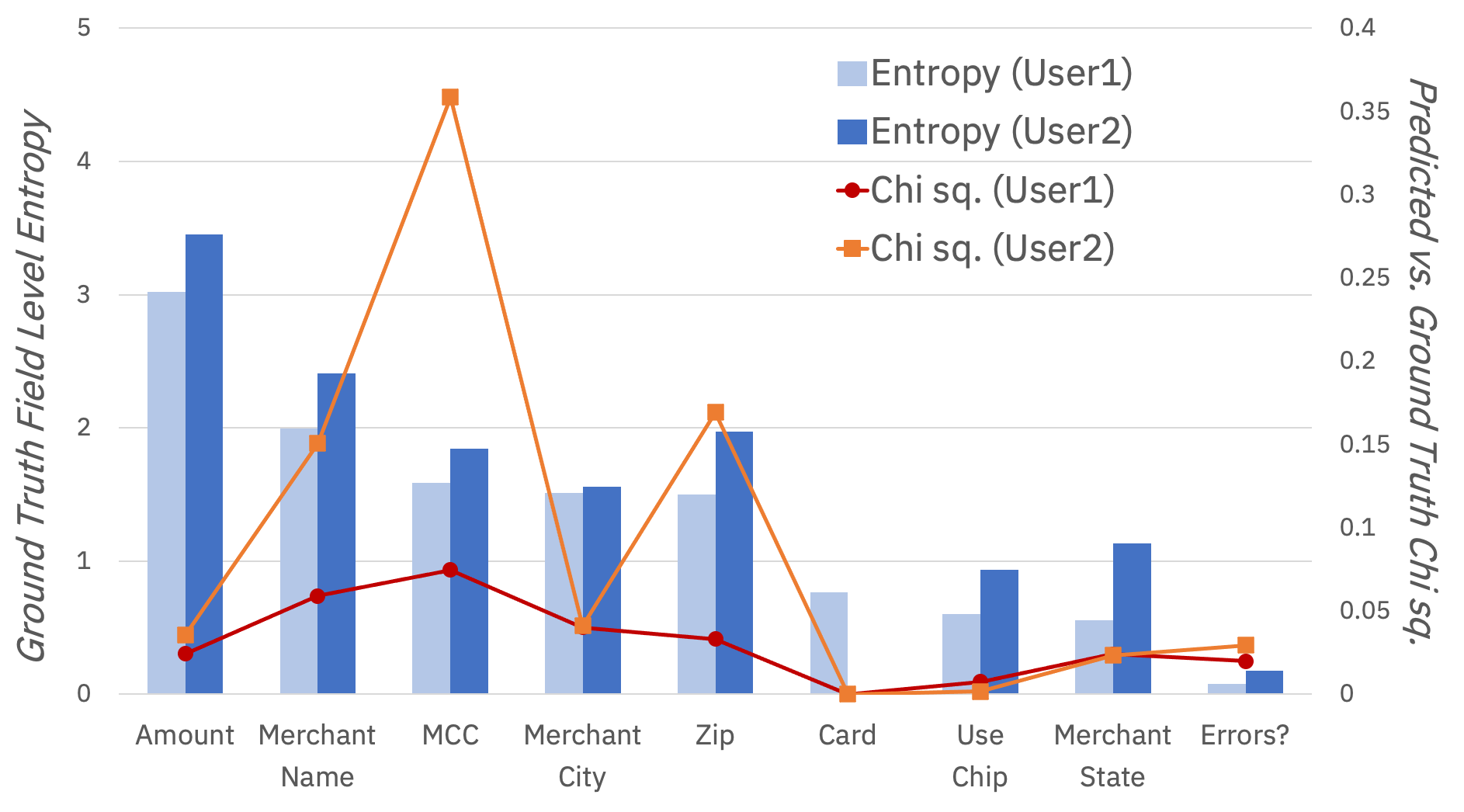}
    \caption{For each column in the tabular data, we compare the generated and ground truth distributions for the user's data rows.
    The entropy of each feature is represented by the bars and displayed on the left vertical axis and $\chi^2$ distance between real and synthetic data distributions is represented by the line and displayed on right vertical axis.}
    \label{fig:gpt_eval}
    \vskip -0.15in
\end{figure}
In \Fig{fig:gpt_eval}, we plot results of this evaluation for the two selected users.
Overall, we see that for both users, their respective TabGPT models are able to generate synthetic distributions that are similar to the ground truth for each feature of the data, even for columns with high entropy, such as \texttt{Amount}.
The TabGPT model for \emph{user 1} produces distributions that are generally closer to ground truth, but for \emph{user 2}, most column distributions also align closely.

While field distribution matching evaluates fidelity of generated data to real data on an aggregate level, this analysis does not capture the sequential nature of the generation.
Hence we use an additional metric that compares two datasets of time series $(t^a_{1,i},\dots t^a_{T,i})_{i=1\dots N}$ and $(t^b_{1,i},\dots t^b_{T,i})_{i=1\dots N}$.
Inspired by the \emph{Fr\'echet  Inception Distance} (FID) \cite{heusel2017gans} used in computer vision and \emph{Fr\'echet InferSent Distance} (FD) \cite{semeniuta2018accurate} in NLP, we use our TabBERT model to embed real and generated sequence to a fixed length vector for each instance  $v^a_i=\text{TabBERT}((t^a_{1,i},\dots t^a_{T,i}))$, and $v^b_i=\text{TabBERT}((t^b_{1,i},\dots t^b_{T,i}))$.
$v^a_i$ is obtained by mean pooling all time-wise embeddings \texttt{SE}$_{t}$ in TabBERT.
Then we compute mean and covariance for each dataset $(\mu_a,\Sigma_a)$ and  $(\mu_b,\Sigma_b)$, respectively.
The FID score is defined as follows:
\begin{align}
    \mathrm{FID}_{a,b}= ||\mu_a-\mu_b||^2_2 + Tr(\Sigma_a+ \Sigma_b - 2 (\Sigma_a\Sigma_b)^{\frac{1}{2}})
\end{align}

\vskip -0.1in
\begin{table}[ht!]
\centering
\begin{tabular}{llcc}
    \toprule
    \multicolumn{2}{l}{\multirow{2}{*}{FID}} & \multicolumn{2}{c}{Real}           \\
    \cmidrule(ll){3-4}    
    \multicolumn{2}{l}{}                     & \multicolumn{1}{r}{\textit{User 1}} &\textit{ User 2} \\
    \cmidrule(ll){2-4}
    \multirow{2}{*}{Real}      & \textit{User 1}       & -                         & 492.92 \\
    \cmidrule(ll){2-4}
    \multirow{2}{*}{GPT-Gen}       & \textit{User 1}       & 22.90                     & 497.68 \\
                               & \textit{User 2}       & 515.94                    & 49.08 \\
    \bottomrule
\end{tabular}
\caption{FID between real and GPT-generated transactions.}
\label{tab:FID}
\vskip -0.1in
\end{table}
FID scores between the transaction datasets for \emph{user 1} and \emph{user 2} are presented in \Tab{tab:FID}.
For the real user data, we see that they have different behaviors, with FID of 492.95.
In contrast, the TabGPT generated data (GPT-Gen) for \emph{user 1} matches the real \emph{user 1} more closely, as can be seen from the relatively low FID score.
The same conclusion holds for GPT-Gen \textit{user 2}.
Interestingly the cross distances between the generated user and the other real user are also maintained.
The combination of the aggregate histogram and FID analyses indicates that TabGPT is able to learn the behavior of each user and to generate realistic synthetic transactions.

\section{Conclusion}
In this paper, we introduce Hierarchical Tabular BERT and Tabular GPT for modeling multivariate times series. We also open-source a synthetic card transactions dataset and the code to reproduce our experiments. This type of modeling for sequential tabular data via transformers is made possible thanks to the quantization of the continuous fields of the tabular data. We show that the representations learned by TabBERT provide consistent performance gains in different downstream tasks. TabBERT features can be used in fraud detection \emph{in lieu} of hand-engineered features as they better capture the intra-dependencies between the fields as well as the temporal dependencies between rows. Finally, we show that TabGPT can reliably synthesise card transactions that can replace real data and alleviate the privacy issues encountered when training off premise or with cloud based solutions \cite{assefa2020generating,efimov2020using}. 


%
%
%


\bibliographystyle{IEEEbib}
\bibliography{refs}

\end{document}